# EVIDENTIAL CONFIRMATION AS TRANSFORMED PROBABILITY

## Benjamin N. Grosof

Computer Science Department
Stanford University
Stanford, California 94305
(415) 497-4368, 497-2273
Arpanet: grosof@sushi

## Abstract

A considerable body of work in AI has been concerned with aggregating measures of confirmatory and disconfirmatory evidence for a common set of propositions. Claiming classical probability to be inadequate or inappropriate, several researchers have gone so far as to invent new formalisms and methods. Some of these have become widely used and theoretically explored [17,15]. We show how to represent two major approaches to evidential confirmation not only in terms of transformed (Bayesian) probability, but also in terms of each other; which:

- unifies two of the leading approaches to confirmation theory, by showing that a revised MYCIN Certainty Factor method [12] is equivalent to a special case of Dempster-Shafer theory;

- gives us a well-understood axiomatic basis, i.e. conditional independence, to interpret previous work on quantitative confirmation theory; in particular,

- provides a new axiomatic analysis and interpretation of Dempster's Rule for an important special case;

- which gives a firmer epistemological basis for acquiring, and for using in decisions, Dempster-Shafer belief functions aggregated via Dempster's Rule;

- substantially resolves the "take-them-or-leave-them" problem of priors: MYCIN had to leave them out, while PROSPECTOR had to have them in;

- recasts some of confirmation theory's advantages in terms of the psychological accessibility of probabilistic information in different (transformed) formats;

- helps to unify the representation of plausible/ inexact/ uncertain reasoning (see also [11]);

- clarifies the place of evidential confirmation in a general scheme for probabilistic reasoning: as concerned with aggregating "parallel" updates or changes in probability; and

- in particular marries evidential confirmation to the strengths of Bayesian, arbitrary-conditional, probability: especially, if-then rules and forward and backward chaining.

## Introduction[1]

---

[1] Limitations of space prevent us from introducing the terminology and definitions of MYCIN, PROSPECTOR, and Dempster-Shafer. We assume the reader is somewhat familiar with them.

A considerable body of work in AI has been concerned with aggregating measures of confirmatory and disconfirmatory evidence for a common set of propositions. On the face of it, this problem may not appear to be compatible with the usual notion of probability. One apparent difficulty is that if we formalize one piece of evidence as:

$$q_1 \leq p(A) \leq r_1$$

and another as:

$$q_2 \leq p(A) \leq r_2$$

then we may have an inconsistency, e.g. if $r_1 < q_2$. Other difficulties also are often cited; however, this paper is not the place to go into them. Claiming classical probability to be inadequate or inappropriate, several researchers have gone so far as to invent new formalisms and methods. Some of these have become widely used and theoretically explored [17,15].

Two of the leading approaches to evidential reasoning are MYCIN-type certainty factors and Dempster-Shafer theory. MYCIN helped to stimulate the interest of the AI community in the problems of probabilistic reasoning in general, and in the concept of confirmation in particular. MYCIN-type certainty factors are widely used, e.g. in PUFF [1] and in current EMYCIN-based expert systems[2]. They also bear substantial similarity to several other scoring schemes used in AI for uncertainty, e.g. INTERNIST's. Dempster-Shafer theory has received much attention in the AI community recently [2,6,13,14,9,7,10,8]. Naturally, there has been interest in trying to relate these two approaches. For example, Jean Gordon and Ted Shortliffe in [9] propose a scheme for a partial mapping of MYCIN Certainty Factors into Dempster-Shafer belief functions. We propose a different mapping which is invertible, i.e. an isomorphism, unlike theirs.

Recent work by David Heckerman [12] providing a probabilistic interpretation for MYCIN's Certainty Factors establishes a new basis for formulating evidential confirmation in terms of probability. He recasts Certainty Factors in such a way as to preserve their formal properties, e.g. as proposed in [17] and later extended in [3], but in addition to satisfy the requirements of a (Bayesian, i.e. conditional) probabilistic interpretation. Moreover, in most cases, the revised certainty factors are numerically quite close to the original versions.

The key technical points of this paper are constituted by the mutual cross-mapping of uncertainty measures **and their aggregation rules** for three different schemes: Heckerman's revised Certainty Factors, PROSPECTOR's likelihood ratios, and Dempster-Shafer belief functions (restricted to the "simple", "chance", case).

We show how to represent two major approaches to evidential confirmation not only in terms of transformed (Bayesian) probability, but also in terms of each other[3]. In particular, the revised MYCIN approach is mapped **exactly** into a special case of Dempster-Shafer theory.

As far as Dempster-Shafer theory goes, the significance is two-fold. Firstly, we discover that both the (revised) Certainty Factor method and the PROSPECTOR method are equivalent to a special case of Dempster-Shafer theory, and thus are in effect subsumed by it.

---

[2] e.g. those developed using the EMYCIN-based technology of Teknowledge, Inc. of Palo Alto, CA.

[3] [11] shows how the more general framework of (Bayesian) conditional *interval* probability can incorporate evidential reasoning; it contains Dempster-Shafer theory as a special case.

Secondly, we have a new analysis of Dempster's Rule, which lies at the heart of any semantic (as well as decision-theoretic) account of Dempster-Shafer theory. We show how by using a non-linear but invertible transformation, we can interpret a special case of Dempster's Rule in terms of a conditional independence assumption. This kind of assumption is something with a history in AI (e.g. PROSPECTOR), and is ubiquitous in the literature on probability theory and applications. Thus we can understand Dempster-Shafer theory, at least in this special case, as doing something fundamentally the same as PROSPECTOR and (Heckerman's revised) CF's. This seems to unify nicely the two leading approaches to evidential confirmation with each other, and with Bayesian probabilistic reasoning, at the deep level of aggregation assumptions, i.e. the semantics of the "operations" on the uncertainty measure "data types".

## CF's Are A Transform of Likelihood Ratios, With Independence Assumption

We show that the revised Certainty Factor method is equivalent to the scheme for combination of evidence used in PROSPECTOR [5]. The revised certainty factors, which are called "$\Delta$s" in [12], correspond via an invertible, non-linear transformation to likelihood ratios, which are called "$\lambda$s" in [5]; the combination rule for revised certainty factors corresponds to the product rule for combining multiple pieces of evidence in [5]. The assumptions underlying each method, [12] and [5], are equivalent: they are the conditional independence of antecedent ("evidence") events given consequent ("hypothesis") events.

## Mapping #1

Let us take the odds-likelihood form of Bayes' Rule as our starting point:

$$\frac{p(H \mid E)}{p(\neg H \mid E)} \bigg/ \frac{p(H)}{p(\neg H)} = \frac{p(E \mid H)}{p(E \mid \neg H)}$$

PROSPECTOR defines the quantity on the right hand side above as a "$\lambda$", here denoted by L(H, E). The combination rule, i.e. aggregation operation, is commutative and associative:

$$L(H, E_1 \wedge E_2) = L(H, E_1) * L(H, E_2)$$

The identity is L = 1. L ranges on $[0, \infty)$.

**N.B.: The underlying assumption here, and thus in each of the equivalent schemes, is that of conditional independence of the $E_i$ given H and given $\neg H$.**

This is used to give posterior odds on H, after applying several pieces of evidence (which might have been several uncertain "if-then" rules, i.e. Bayesian, conditional probabilities, all concluding the same proposition, H), from prior odds on H:

$$\frac{p(H \mid E_1 \wedge \ldots \wedge E_n)}{p(\neg H \mid E_1 \wedge \ldots \wedge E_n)} = L(H, E_1 \wedge \ldots \wedge E_n) * \frac{p(H)}{p(\neg H)}$$

Heckerman's revised certainty factor, which he calls "$\Delta$", is here denoted by D(H, E). It is arrived at by the invertible transformation[4]:

---

[4] This author originated the mapping of Heckerman's scheme to PROSPECTOR's; he discusses it in his paper [12]; see his acknowledgements. We are coordinating writing the two papers so that they will be, hopefully, coherent yet complementary.

$$D = \frac{L-1}{L+1}$$

$$L = \frac{1+D}{1-D}$$

The combination rule (aggregation operation) is (naturally, commutative and associative):

$$D(H, E_1 \wedge E_2) = \frac{D(H, E_1) + D(H, E_2)}{1 + D(H, E_1) * D(H, E_2)}$$

The identity is $D = 0$. $D$ ranges on $[-1, +1]$.

## Combining CF's Is Multiplying Likelihood Ratios Is Dempster's Rule

We show furthermore that both the revised Certainty Factor method (D) and the PROSPECTOR method (L) are equivalent to a special case of Dempster-Shafer theory [15,4]. Via another invertible transformation, we show a correspondence between $\Delta s$ (D), and thus $\lambda s$ (L), and belief measures (B, below) on a $[0, 1]$ scale. We treat this belief measure as a Dempster-Shafer belief function. The combination of $\Delta s$, and of $\lambda s$, then corresponds exactly to the application of Dempster's Rule to ordinary single-valued probabilities. The conditional independence assumption underlying the PROSPECTOR updating rule is thus discovered to be equivalent to a special case of Dempster's Rule.

## Mapping # 2

Dempster-Shafer Belief we will denote here by B(H, E), where E represents the index of the belief function, which we can regard as the evidential "source" of the belief function. Dempster-Shafer belief is arrived at by another, invertible transformation:

$$B = \frac{L}{L+1}$$

$$L = \frac{B}{1-B}$$

and, equivalently:

$$B = \frac{1+D}{2}$$

$$D = 2*B - 1$$

The combination rule (aggregation operation) is just Dempster's Rule, here specialized to point-valued ("chance" in Shafer's terminology") belief:

$$B(H, E_1 \wedge E_2) = \frac{B(H, E_1) * B(H, E_2)}{B(H, E_1) * B(H, E_2) + [1 - B(H, E_1)] * [1 - B(H, E_2)]}$$

The identity[5] is B = 0.5. B ranges on [0, 1].

Note that here $[1 - B(H, E_i)] = B(\neg H, E_i)$, since we are in the special "chance" case. More precisely, $B(., E_i)$[6] corresponds to a class of Dempster-Shafer belief functions: those which are the "extensions" (onto more fine-grained frames of discernment) of the belief function $Bel_{i0}(.)$, where $Bel_{i0}$ is defined on the frame of discernment $\{H, \neg H\}$ as equivalent to the "mass" function $Mass_{i0}$ (on the same frame of discernment):

$$Mass_{i0}(H) = B(H, E_i)$$
$$Mass_{i0}(\neg H) = B(\neg H, E_i) = 1 - B(H, E_i)$$

We will call $B(., E_i)$ an "extended, simple, chance" belief function. "Simple" here means that the "focus", i.e. set of propositions with non-zero mass, is minimal in size[7]: i.e. has only two members. "Chance" refers to the focus being a subset of the frame of discernement, i.e. of the set of primitive propositions[8].

## Viewing Evidence As Updates Versus As Priors

We can think of all three schemes as manipulating **changes in probability** (which we can think of as updates), while relying on a null "prior" of sorts: the identity for the combination rule. While in PROSPECTOR, priors were specified separately, in terms of straight probability (in odds form, actually) instead of as an L, in the Certainty Factor and Dempster-Shafer approaches, all evidence is given equal status. However, we see that a substantive "prior" in the PROSPECTOR approach, e.g.

$$odds(H) = \frac{p(H)}{p(\neg H)} = X$$

is just equivalent to a piece of evidence

```
L(H , PriorInfo)  =  X
```

aggregated with the vacuous "system prior", i.e. identity:

```
L(H , {})  =  1 .
```

**Thus the requirement for priors in PROSPECTOR is no handicap, relative to Certainty Factors and Dempster-Shafer.** Indeed, those who avoid specifying priors in the latter schemes are in fact making an assumption by relying implicitly on the vacuous prior.

Similarly, we can encode arbitrary priors into Certainty Factors. If our prior on H is:

```
p(H | PriorInfo)  =  Y
```

then we can represent it as just another certainty factor:

---

[5] If you thought the identity for Dempster's Rule is the bounded interval [0, 1] don't worry: this is a special case, remember.

[6] Our notation F(.) means the function F as a whole, i.e. ($\lambda x. F(x)$) in lambda-calculus notation.

[7] without the belief function being trivial, i.e. representing certainty

[8] The terminology of Dempster-Shafer theory used here is drawn from Shafer's [15].

$$D(H, PriorInfo) = 2*Y - 1$$

Thus each scheme, because of its commutative and associative combining rule, can be regarded as treating all pieces of evidence (updates) on equal terms: everything and nothing is a "prior".

## Synthesis

It makes sense that the mapping of (revised) Certainty Factors (D) is to the "one-number" special case (as opposed to the "two-number" general case, i.e. bounded intervals) of Dempster-Shafer belief.

It makes sense that (revised) Certainty Factors (D) are mapped into the the special case of Dempster-Shafer belief in which bounds meet. After all, Certainty Factors represent the uncertainty of a proposition with only one number. Much of the historical modification of the Certainty Factor approach (e.g.[3]), and much of its ad-hoc feel and formal difficulties, have revolved around the summarizing of the evidence in a single number rather than a pair. (Originally, there were MB and MD, "measure of belief" and "measure of disbelief", repectively.) Our mapping helps to show in what way full Dempster-Shafer theory is richer and more powerful than Certainty Factors. This helps to justify and rationalize the AI community's (e.g. Shortliffe's) interest in, and attraction to, Dempster-Shafer theory.

Each of the three schemes seems to be useful in different ways. L is particularly helpful in characterizing the assumption underlying the combination rule[9], and thus in providing a sound probabilistic interpretation for D, as well as an alternative probabilistic interpretation for B. D has been found to be psychologically useful in MYCIN and EMYCIN work, e.g. especially for knowledge acquisition. D has a nice symmetry, represented by numerical negation, between confirmatory and disconfirmatory evidence. Now we have a rigorous and understandable axiomatic "semantics" for the meaning of the numbers in D. Finally, B is useful for understanding D and L to be subsumed by Dempster-Shafer theory.

L also provides a link to conditional probabilities and thus to applying *chains* of uncertain "if-then" rules. We can regard "evidential reasoning" or "confirmation theory", in the sense of Certainty Factors and Dempster-Shafer theory, as being concerned with how to combine, i.e. aggregate, several probabilistic conclusions "in parallel", i.e. which all bear on the same proposition or set of propositions. Each of these probabilistic conclusions may be rather primitive, i.e. an irreducible estimate from some opaque source, or it may be arrived at by an extensive chain of conditional probabilistic reasoning. See [11] for more discussion.

## Conclusions

Understanding evidential confirmation in terms of transformed (single-valued, Bayesian) probability:

- unifies two of the leading approaches to confirmation theory, by showing that the (revised) MYCIN Certainty Factor method is equivalent to a special case of Dempster-Shafer theory;

- gives us a well-understood axiomatic basis, i.e. conditional independence, to interpret previous work on quantitative confirmation theory; in particular,

- provides a new axiomatic analysis and interpretation of Dempster's Rule for an important special case;

---

[9] G, defined as (ln L), is even more conceptually elegant; it has an *additive* aggregation rule. Indeed, Heckerman originally conceived the aggregation rule for D in terms of G.

- which gives a firmer epistemological and "canonical" ([16]) basis for acquiring, and for using in decisions, Dempster-Shafer belief functions aggregated via Dempster's Rule;

- substantially resolves the "take-them-or-leave-them" problem of priors: MYCIN had to leave them out, while PROSPECTOR had to have them in;

- recasts some of confirmation theory's advantages in terms of the psychological accessibility of probabilistic information in different (transformed) formats;

- helps to unify the representation of plausible/ inexact/ uncertain reasoning (see also [11]);

- clarifies the place of evidential confirmation in a general scheme for probabilistic reasoning: as concerned with aggregating "parallel" updates or changes in probability; and

- in particular marries evidential confirmation to the strengths of Bayesian, arbitrary-conditional, probability: especially, if-then rules and forward and backward chaining.

## ACKNOWLEDGEMENTS

This work was supported by the author's National Science Foundation Graduate Fellowship, by the KSL-MRS Project at Stanford University, and by the following grants: DARPA N00039-83-C-0136; ONR N00014-81-K-0004. The author thanks Michael Georgeff, Eric Horvitz, and, especially, David Heckerman, for valuable discussions. Thanks to Nils Nilsson and Matt Ginsberg for their comments on an earlier draft of this paper. Thanks also to Richard Duda, Peter Hart, and Nils Nilsson for their interest, and their help in providing the unpublished history of the development of the PROSPECTOR method.